\documentclass[conference]{IEEEform/IEEEtran}

\usepackage{cite}
\usepackage{url}
\usepackage{amsmath,amssymb,amsfonts}
\usepackage{algorithmic}
\usepackage{graphicx}
\usepackage{textcomp}
\usepackage{xcolor}
\def\BibTeX{{\rm B\kern-.05em{\sc i\kern-.025em b}\kern-.08em
    T\kern-.1667em\lower.7ex\hbox{E}\kern-.125emX}}

\begin{document}

\title{\title{Integration of Convolutional Neural Networks in Mobile Applications}
}

\author{\IEEEauthorblockN{Roger Creus Castanyer}
\IEEEauthorblockA{\textit{Universitat Politècnica de Catalunya}\\
Barcelona, Spain \\
roger.creus@estudiantat.upc.edu}
\and
\IEEEauthorblockN{Silverio Martínez-Fernández}
\IEEEauthorblockA{
\textit{Universitat Politècnica de Catalunya}\\
Barcelona, Spain \\
silverio.martinez@upc.edu}
\and
\IEEEauthorblockN{Xavier Franch}
\IEEEauthorblockA{\textit{Universitat Politècnica de Catalunya}\\
Barcelona, Spain \\
franch@essi.upc.edu}
}

\maketitle
\begin{abstract}
When building Deep Learning (DL) models, data scientists and software engineers manage the trade-off between their accuracy, or any other suitable success criteria, and their complexity. In an environment with high computational power, a common practice is making the models go deeper by designing more sophisticated architectures. However, in the context of mobile devices, which possess less computational power, keeping complexity under control is a must. In this paper, we study the performance of a system that integrates a DL model as a trade-off between the accuracy and the complexity. At the same time, we relate the complexity to the efficiency of the system. With this, we present a practical study that aims to explore the challenges met when optimizing the performance of DL models becomes a requirement. Concretely, we aim to identify: {\it (i)} the most concerning challenges when deploying DL-based software in mobile applications; and {\it (ii)} the path for optimizing the performance trade-off. We obtain results that verify many of the identified challenges in the related work such as the availability of frameworks and the software-data dependency. We provide a documentation of our experience when facing the identified challenges together with the discussion of possible solutions to them. Additionally, we implement a solution to the sustainability of the DL models when deployed in order to reduce the severity of other identified challenges. Moreover, we relate the performance trade-off to a new defined challenge featuring the impact of the complexity in the obtained accuracy. Finally, we discuss and motivate future work that aims to provide solutions to the more open challenges found.
\end{abstract}

\section{Introduction}
Convolutional Neural Networks (CNN) have become a widely used architecture since they have proven to be an accurate universal function estimator in many domains, such as Image Classification and Natural Language Processing \cite{CNN-lecun}. As a particular type of Neural Network (NN), a CNN consists of a collection of parameters, which represent weights and biases, that are tuned depending on the problem that is faced. These parameters, put together after long training sessions, have proven to be able to solve  complex problems even outperforming humans \cite{GO-deepmind}. The tuning is performed automatically and the values that the parameters take are outside of the human comprehension, turning the models into black boxes. 

The use of CNNs in custom applications demands high computational power, since there is no physical limitation to the complexity of the designed models. Simple augmentation of the size of the collection of trainable parameters has proven to work as long as new data keeps being fed to the model \cite{hestness2017deep}. Furthermore, following the exponential evolution of the availability of computational power, the tendency of solving complex tasks with complex models is increasing \cite{anthes2017artificial}. Language models are good examples of this, e.g. Mann et al. with the description of a NN with 175 billion parameters \cite{gpt3}.

According to this tendency, the most common deployment platforms used historically for DL-based software are servers that enable cloud computing. For the purpose of making use of a trained NN, a data pipeline is developed manually so that it links the pre-processing steps, computation of results, post-processing and eventually the display of the predictions. When all these steps are implemented in a server/cloud, the requirement of an internet connection for using the model is a must. However, with the growth of the mobile applications market, approaches on the integration of NNs in lightweight devices are becoming popular given their ability to provide great services to the user. Within this setting, DL-based software consists of software-reliant systems that \textbf{include data and components} that implement algorithms mimicking learning and problem solving \cite{arpteg2018software}.

In this work, we aim to perform an exploratory and descriptive analysis on: {\it (i)} what are the current challenges regarding the deployment of DL-based software as mobile applications; and {\it (ii)} how can complexity be controlled while keeping a good level of accuracy. The subject is analysed by means of a practical study in which we perform all the required cycles from data acquisition, DL modelling, classification and application, and operation in real context. In this way, we study the relation between the DL-based software development and the accuracy when operating with it in the production settings \cite{9238323}. Ensuring the generalization of the DL-based software in the operation environment is key and can be especially concerning in safety-critical applications \cite{martinez-fernandez}.

Our work is performed over the German Traffic Sign Recognition Benchmark dataset \cite{Benchmarking-tsr}, created in order to standardize the traffic sign recognition literature. In the DL literature, this faced task is an instance of the Image Classification problem \cite{doi:10.1080/01431160600746456}. Each image in the data is from one and only one class so the models will be trained using classification accuracy as the success criteria. CNNs are the standard approach to Image Classification problems for their ability to fit in image processing. The CNN architecture has already shown that can achieve very good results in the literature but also for this specific dataset \cite{li2014medical} \cite{Benchmarking-tsr}.

This work has four main contributions: 
\begin{itemize}
    \item Show experiences of the end-to-end DL lifecycle from a practical study on traffic sign recognition.
    \item  Discuss the challenges encountered in the deployment of CNNs.
    \item Discuss the trade-offs between accuracy and complexity in environments with limited computational power (e.g., mobile applications).
    \item An open science package of the developed DL-based software freely available on GitHub, following the "Cookiecutter Data Science" project structure \footnote{https://drivendata.github.io/cookiecutter-data-science/} to foster correctness and reproducibility.
\end{itemize} 

The document is structured as follows. In Section 2 we describe Related Work on both the challenges of integrating DL models in mobile applications and the focuses on optimizing the performance trade-off. In Section 3 we describe the Study Goal and Research Questions. In Section 4 we describe the study design and the end-to-end DL software lifecycle. In Section 5 we show the results of practical study that allow to answer the Research Questions. In Section 6 we discuss what is strictly analysed in our study and what specific parts of the whole topic are addressed.
To end with, in Section 7 we draw conclusions from the research performed and motivate future work.

\section{Related Work}
In this section, we respectively describe the challenges of deploying CNNs in DL-based software, and related work on the optimization of the performance trade-off in systems with limited computation power.

\subsection{Challenges in deployment}\label{sec:sota-challenges}
Regarding the recent interests in the deployment of DL-based software, Chen et al. state that mobile applications come with the strongest popularity trend over other platforms like servers/clouds or browsers for integrating DL models \cite{deployment-DL}. They also show that this increasing trend in the deployment of DL models in mobile applications comes with an inverse proportional relation with the knowledge that the community has about the subject. This way, they demonstrate that the deployment of DL-based software becomes the most challenging part in the life cycle of DL, and that this effect is even more critical when the target are mobile devices.

When analysing the challenges of deployment of DL-based software, several studies focus on the quality attributes that are more relevant to this type of software. The importance of each quality attribute varies depending on the application but it is always the case that DL-based software implies taking care of measures that might not be considered when building traditional software systems. Remarkably, DL-based software effectiveness strongly depends on the structure of the data.

Indeed, Pons and Ozkaya clearly state that, compared to software systems that do not integrate DL-based components, the deployment of DL-based software increases the risk in many quality attributes \cite{quality-AI}. They identify Data centricity as the cause that affects the robustness, security, privacy and sustainability of these systems. Additionally, they analyse the methodology for architecting in Artificial Intelligence (AI) engineering. With the focus put on the software-data dependency, Ozkaya makes the difference between the development of software systems and DL-based software in the processes of building and sustaining \cite{engineering-AI}. Also, Lwakatare et al. provide a list of challenges and solutions regarding the life cycle of DL-based software within industry settings. The challenges and solutions are synthesized into four quality attributes: adaptability (e.g. unstable data dependencies and quality problems), scalability (e.g. balancing efficiency and effectiveness in DL workflow), safety (e.g. explainability of DL models) and privacy (e.g. difficult data exploration in private datasets) \cite{large-scale-AI}.

The identified challenges in the above related work that are objects of study in our work are: 
\begin{itemize}
    \item (C1) \textbf{Frameworks in early stage} of its development to support DL-based mobile applications.
    \item (C2) \textbf{Software--Data dependency}.
    \item (C3) \textbf{Explainability} of the models.
    \item (C4) \textbf{Sustainability} of the models when deployed.
\end{itemize}
   
\subsection{In the quest for optimizing the performance}
In the pursuit of optimizing the performance of DL-based software, there exist several key points that affect the overall system implementation. Important bottlenecks that involve limitations to the efficiency are found in both the modelling stage and in the usage of different frameworks to deploy the models. On the one hand, recent contributions \cite{mobile-nets} show an optimized CNN architecture for enabling its usage in devices with less energy capacity. This allows to reduce the model's complexity (e.g. storage weight, computing power) while keeping the desired accuracy, hence optimizing the performance. In this case, the increase in the performance is due to the increase in the efficiency. 
On the other hand, the availability of different frameworks supporting the implementation of DL-based software in all its stages has a strong influence on the accuracy of the integrated DL models, as shown in \cite{empirical-frameworks}. 

Compared to the aforementioned studies, in our work we focus on analysing the challenges when deploying a specific type of NN, namely CNN, into mobile devices. Furthermore, we do so by means of a practical study. We relate the challenges found with the analysis of the performance, defined as the trade-off between two quality attributes: accuracy and complexity. We provide a comparative analysis between different configurations for each of the models in order to gather evidence of the implications of the accuracy and complexity. Moreover, the sustainability of the models once deployed is also studied to enhance the capabilities of the applications during the DL-based software lifecycle. Our work is developed under the PyTorch framework and Android operating system.

\section{Research Questions}
In this Section, we define the Research Questions (RQs), following the GQM guidelines \cite{GQM}.

\begin{itemize}
\item RQ1 - What are the challenges of the creation and integration of complex CNNs in DL-based mobile applications?

\item RQ2 - What criteria are needed to reason about the trade-off between accuracy and complexity of DL models in mobile applications?

\end{itemize}

The motivation of RQ1 within this project is to study the \textbf{challenges of the creation, training and integration of a CNN in a mobile application}, while identifying, exploring and documenting as many challenges as possible. The obtained conclusions aim to be generic for anyone whose desire is deploying complex models in quotidian software applications.

Furthermore, RQ2 motivates a \textbf{comparative analysis between different solutions to the problem of optimizing the performance trade-off}, since balancing the accuracy of the models and their complexity can be studied from different viewpoints (e.g. design of optimized operations or limitation of architecture's complexity).

\section{Study design}\label{study-design}
In this section we describe the proposed pipeline for achieving a functional implementation of the traffic sign recognition mobile application. 

There are four main cycles in the DL-based software lifecycle \cite{9238323}. First, the \textit{Data Management} which consists of the collection and processing of raw data. Second, the \textit{DL Modelling} which consists of adjusting different models to obtain the best-performing possible solution. Third, the \textit{Development} of the environment (mobile application in our work) and the integration of the DL model in it. Fourth, the \textit{DL-based System Operation} with the application that integrates the DL model. This last phase enables sustainability of the model and experiments the performance trade-off within the production settings. 

\begin{figure*}[t]
    \centering
    \includegraphics[width=\textwidth]{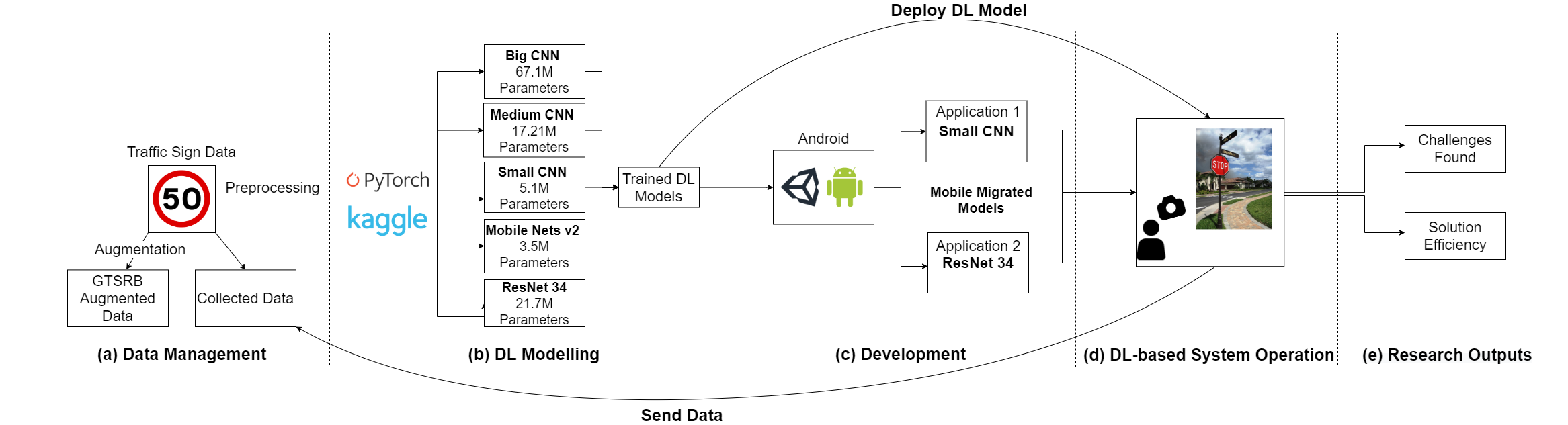}
    \caption{Software development lifecycle in our DL-based software}
    \label{global_overview}
\end{figure*}

A global overview of the study design and DL-based software lifecycle is shown in Figure \ref{global_overview}. The goals required over the system are shown in the following list:
\begin{enumerate}
    \item Use of the device integrated camera
    \item Real-time response for a call to the model
    \item High accuracy in traffic sign recognition
    \item Data augmentation by the community
    \item Integration of updated models 
\end{enumerate}
We implement these goals through iterations along the aforementioned DL-based software lifecycle phases. We also provide a solution for the \textit{Send Data} operation in Figure \ref{global_overview}. This allows the possibility that the user collects the generated data at run-time during the DL-based software system operation, which is then manually annotated and used for re-training the models on the performant computer. Then, this updated models can be integrated again for inference in the mobile application.

\subsection{Platforms and technologies used}
For joining the DL modelling and the development of DL-based software, there exist recent DL frameworks, all of them under continuous development. These frameworks offer up to two different approaches to model transferring operations. On the one hand, \emph{model conversion}. CoreML for iOS software and TensorFlow-Light for Android arguably are the most famous examples that offer this capability. On the other hand, \emph{model export} using the Open Neural Network Exchange (.ONNX) file format, which was presented in 2017 by Microsoft and Facebook. This file format allows to store serialized versions of trained networks in compressed files. 

Since the .ONNX file format is widely used in the current state of the art, it is defined as a key part part of the proposed project pipeline for the practical study.

With all this, the focus is put on finding two different platforms that allow development-sided and operation-sided implementation of both the DL-component and the application that integrates it. The development-sided platform has to allow training and writing of a CNN into a .ONNX file. The operation-sided one has to be capable of reading and using the model and also of supporting the use of the camera in a mobile device. In this work, the development-sided platform is set to be PyTorch and the operation-sided platform is set to be Unity. We make use of the Unity Barracuda libraries for connecting the two platforms. Also, as observed in Figure \ref{global_overview} the models built on PyTorch are trained on the Kaggle GPU, which is available freely for 40 hours a week.
 
Moreover, it is also seen in Figure \ref{global_overview} that only two of the five models that are trained are later deployed to the application. For an appropriate analysis of the performance of these two models when deployed, we built two applications that only vary in the integrated models.
 
\subsection{The Dataset}
The GTSRB traffic sign data consists of 144,769 labelled images of 2,416 traffic sign instances of 70 classes belonging to a portion of the real traffic signs that can be found in the roads of Germany. However, after following several criteria for guaranteeing the quality of data, explained in \cite{Benchmarking-tsr}, the final dataset consists of a collection of 51,840 images from 43 classes fulfilling: {\it (i)} images are of sizes from 15×15 to 222×193 pixels, stored in PPM format; {\it (ii)} it contains the definitions of the region of interest with a separating margin of the size equal to the 10\% of the pixels; {\it (iii)} it records the temporal order of the creation of the images; {\it (iv)} it provides post-processed features like HOG descriptors, Haar-like features and Color Histograms. No usage of the post-processed features is made for the purpose of the deployment of a full-stack convolutional-based architecture for solving the image classification problem. 
\subsubsection{The Pre-processing}
We applied several pre-processing steps that can be considered critical in the context of image processing model training: {\it (i)} the images are resized to a fixed and squared size; {\it (ii)} a center crop is applied to the images; {\it (iii)} The images are encoded as tensors of the form \textit{[B, C, H, W]}, where \textit{B} is the batch size, \textit{C} is the number of channels, \textit{H} is height and \textit{W} is width. The values of these tensors are normalized into the 0-1 range; {\it (iv)} the tensors are standardized in the \textit{H}x\textit{W} channels with the pre-computed means and standard deviations of the 3 RGB channels of the images in the dataset.

\subsubsection{The Augmentation}
For the purpose of supporting a comparative analysis of the performance of different models, we implemented two variants for applying the technique of data augmentation to the pre-processed dataset: {\it (i)} rotation in 90º degrees plus the vertical and horizontal flips to the images; {\it (ii)} addition of manually tagged images by the users of the application during the system operation phase. 

\subsection{The DL/CNN models}
The models that are applied for solving the traffic sign recognition task are the standard CNN, the MobileNet v2 (MBv2) \cite{mobile-nets} and the ResNet34 (RN34) \cite{resnet}. For the first one, we test and evaluate different configurations in order to choose an optimal one that balances accuracy and complexity. For the last two, we test and compare two different ways of training the model: pre-training with feature extraction and fine-tuning, and full-training from scratch. The CNN works with standard convolutional blocks that consist of convolutions followed by batch normalization, an activation function, a max-pooling layer and a dropout layer. It can learn representations of images from scratch which are then passed to a classifier that is built on top of the convolutional blocks. The MBv2 is similar but works with separable depthwise convolutions. It is a form of factorized convolutions which factorize a standard convolution into a depthwise one and a 1x1 one called a pointwise convolution. The depthwise convolution applies a single filter to each input channel and the pointwise convolution then applies a 1x1 convolution to combine the outputs of the depthwise convolution \cite{mobile-nets}. Finally, the RN34 is a more complex CNN that has been successfully applied in the image classification domain and has become a benchmark model for related tasks in the DL literature \cite{resnet}. Both the MBv2 and the RN34 used in the practical study are downloaded from PyTorch and  are pre-trained on ImageNet data \cite{deng2009imagenet}. 

The results on the CNN models are shown in Table \ref{tab:cnn-results}, where we compare the number of Convolutional Layers, the sizes of the two implemented Fully Connected layers (FC1 and FC2), the total number of parameters, and also the accuracy achieved in the test set with 10 epochs of training. Furthermore, we write down if the augmented dataset is used, the fixed image sizes chosen and the model weight in MegaBytes (MB). Note that the input size of FC1 is equal to the number of output features of the convolutional blocks. Also note that the output size of FC1 is the input size of FC2 and the output size of FC2 is 43, which is the number of unique classes of traffic signs in the data. The resulting optimal configuration from Table \ref{tab:cnn-results} is promising because of the little complexity, the small storage weight and the great performance shown.

\begin{table*}[t]
\begin{center}
\caption{\label{tab:cnn-results}Performance analysis of the different CNN configurations.}
\scalebox{1}{%
    \begin{tabular}{|c|c|r|c|c|c|r|r|r|}
        \hline
        \textbf{Data Aug.} & \textbf{Image Size} & \textbf{Batch Size} & \textbf{Conv. Layers} & \textbf{FC1 input} & \textbf{FC2 input} 
        & \textbf{Num. Parameters} & \textbf{Test accuracy} &  \textbf{Model Weight} \\
        \hline
         NO & 256x256 & 256 & 5 & 8x8x256 & 4x4x64 & 17.21M & 88.02\% & 65.68MB \\
         NO & 512x512 & 64 & 6 & 8x8x512 & 4x4x128 & 67.13M & 95.00\% & 259.70MB \\
         YES & 256x256 & 64 & 5 & 8x8x256 & 4x4x64 & 17.21M & 89.31\% & 65.68MB \\
         YES & 512x512 & 64 & 6 & 8x8x512 & 4x4x128 & 67.13M & \textbf{95.52\%} & 259.70MB \\
         NO & 256x256 & 64 & 6 & 4x4x512 & 2x2x128 & \textbf{5.10M} & \textbf{95.00\%} & \textbf{22.11MB} \\
         \hline
    \end{tabular}}
\end{center}
\end{table*}

As it can be seen in Table \ref{tab:cnn-results}, no increase in performance appears when the augmented set of data is used. This might be because very fast training is happening, thanks to the non-saturating ReLU function in combination of the convolutional operations plus the use of minibatches. More importantly, it can be seen that the highest tier classification accuracy can be achieved with the lightest of the tested configurations.

Regarding the MBv2, the experiments do not change its architecture but only its training methodology. Three techniques are distinguished: Feature Extraction (FE), which uses a pre-trained version of the model and just tunes the built classifier on top of it, training its weights and biases. Fine-Tuning (FT), which also uses the pre-trained version of the model but tunes the weights of all its layers. And Training from Scratch (TfS), which loads the architecture with a random initialization and fully trains it. The results on the training stage with the MBv2 for 10 epochs are presented in Table \ref{tab:mobilenets-test}.

\begin{table}[h!]
\centering
\caption{Performance analysis of the Mobile Net v2}
\label{tab:mobilenets-test}
\scalebox{1}{%
    \begin{tabular}{|c|c|r|}
        \hline
         \textbf{Data Aug.} & \textbf{Training method} & 
         \textbf{Test accuracy} \\
         \hline
         NO & FE & 4.00\%  \\
         NO & FT & 4.00\%  \\
         NO & TfS & 4.00\% \\
         YES & FE & 68.00\%  \\
         YES & FT & \textbf{95.20\%}  \\
         YES & TfS & 92.50\%  \\
         \hline
    \end{tabular}
}
\end{table}

An important drawback of the MBv2 is spotted in Table \ref{tab:mobilenets-test}, where it can be seen that without the augmented set of data the network is not capable of learning on the traffic sign recognition task. Anyway, when the augmented set of data is used it can be seen that the same performance as the best CNN is achieved with a much smaller number of parameters, showing that this architecture performs very efficiently. 

Finally, the results on the benchmark RN34 is shown in Table \ref{tab:resnet-test}. Training experiments follow the same logic as with the MBv2 but without trying the augmented set of data, since it is rapidly seen that it is not needed for this architecture.

\begin{table}[h!]
\centering
\caption{Performance analysis of the RN34}\label{tab:resnet-test}
\scalebox{1}{%
    \begin{tabular}{|c|c|r|}
        \hline
         \textbf{Data Aug.} & \textbf{Training method} & \textbf{Test accuracy}  \\
         \hline
         NO & FE & 20.00\%  \\
         NO & FT & \textbf{97.80\%}   \\
         NO & TfS & 94.10\%   \\
        \hline
    \end{tabular}
}
\end{table}

As it is seen in Table \ref{tab:resnet-test} the training method of FT gets to the highest seen accuracy in the test split of the training set during the practical study. The output model of this training stage is the one deployed in the application.

\subsection{The DL-based software}
\begin{figure*}[h]
    \centering
    \scalebox{0.65}{%
    \includegraphics[width=\linewidth]{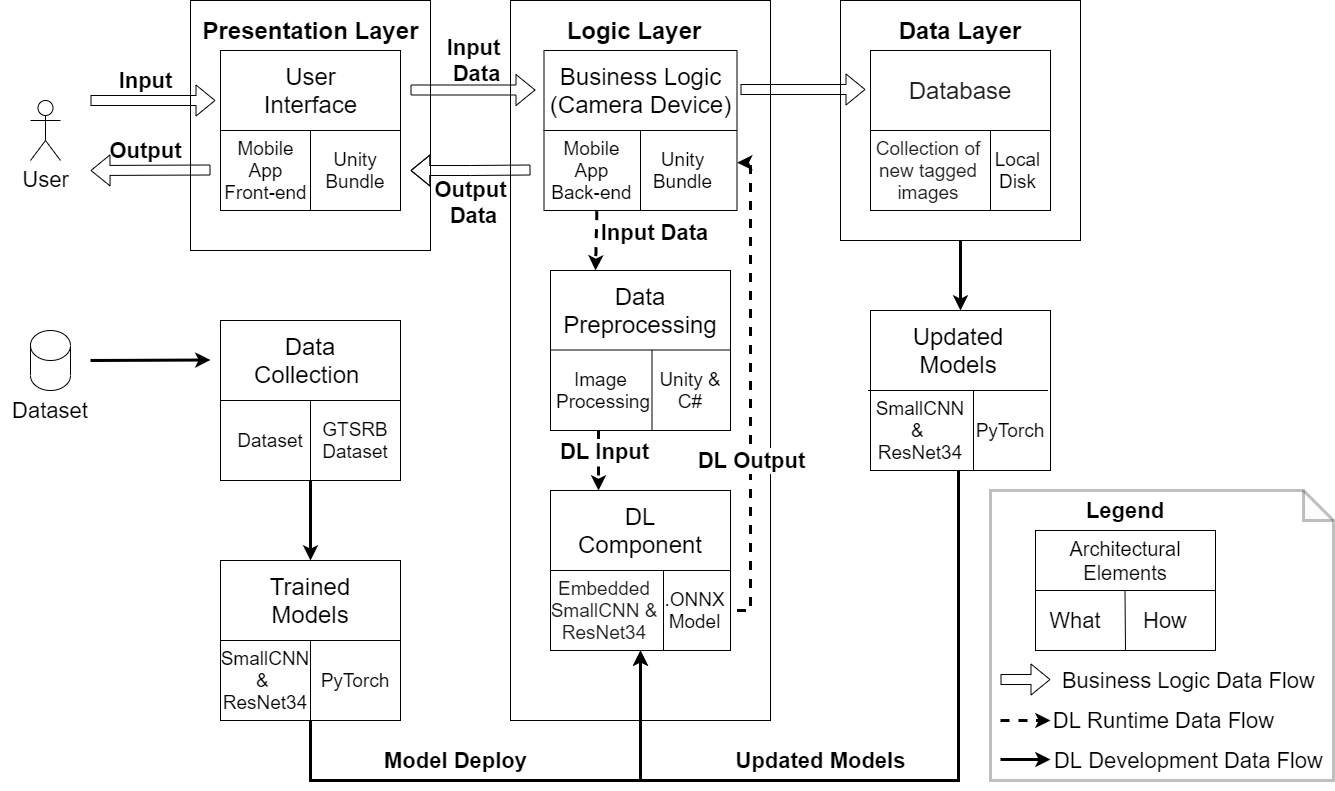}
    }
    \caption{DL-based Software System Operation Architecture}
    \label{app_DL}
\end{figure*}

The architecture of the DL-based software system operation is shown in Figure \ref{app_DL}, following design patterns from \cite{washizaki2020machine}. We build two applications that have the same architecture but only differ in the DL model they integrate. The application that loads the SmallCNN has a total weight of 68.84MB and the one that loads the RN34 of 131MB. The baseline weight of the application bundle generated by Unity3D is of 46.73MB without the models.

The two built applications implement the goals 1,2 and 3 defined in Section \ref{study-design}. However, for ensuring the sustainability of the models when deployed, and for satisfying the applications goals 4 and 5, an external server/computer is needed for performing the re-training of the model with the new tagged data added by the community. This implementation still lets the user obtain recognition predictions without external calls. The user has this option because the application client will only make use of the server/computer for downloading new updates on the model and for manually submit new tagged data in a personally controlled way, whenever it is desired. We consider allowing the user to annotate the data that it generates at run-time an efficient solution to the sustainability of the model when deployed. This could be done either by verifying correct model recognition predictions or by correcting wrong ones. However, the architecture that we implement in this practical study makes use of the mobile device storage to save the taken pictures, which are then manually annotated and sent to the local machine, where they are fed to the model again in the Kaggle GPU. This is not a scalable implementation but releasable for research purposes.

\subsection{Threats to validity}
Regarding construct validity
, we build two DL models following a small CNN and ResNet34 (RN34) \cite{resnet}, in order to mitigate mono-operation bias. As for conclusion validity
, due to the exploratory nature of the study we do not execute statistical tests. Regarding internal validity
, the DL models are executed in the same context (e.g., computational power, mobile technologies), mitigating that the efficiency results are caused accidentally rather than by the DL models themselves. Finally, regarding external validity
, our results are tied to the context of traffic sign recognition in mobile applications. Furthermore, the analysis of the performance in the operation-sided environment is carried out in a mobile device with the Qualcomm Adreno 618 GPU running over Android, which determines the extent to which the results obtained that regard the complexity can be generalized.

\section{Results}
In this section we present and discuss the results of our practical study for each RQ. 

\subsection{RQ1: Challenges of the creation and integration of complex CNNs in DL-based mobile applications}
For answering the RQ1, we analyze both the identified challenges in the state-of-the-art (see Section \ref{sec:sota-challenges}) and our results. This analysis is shown in Table \ref{results-challenges}, where: \textbf{v} indicates that the challenge has been verified to be real and concerning; \textbf{±} indicates that the challenge has been identified but not faced in the practical study; \textbf{new} defines newly spotted challenges up to our knowledge emerging from our study.

\begin{table}[h]
\begin{center}
\caption{\label{results-challenges}Diagnoses of identified challenges}
\scalebox{1.15}{
        \begin{tabular}{|c|c|}
        \hline
        \textbf{Challenges} & \textbf{Diagnostic}\\
        \hline
        Frameworks in early stage (C1) & v \\
        Software-Data dependency (C2) & v \\
        Explainability (C3) & ± \\
        Sustainability (C4) & v \\
        Software dependencies (C5) & new \\
        Model performance (C6) & new \\
        \hline
    \end{tabular}
}
\end{center}
\end{table}

We verify C1 because we find few alternatives to develop DL-based mobile applications. Furthermore, we encounter difficulties in the support of DL architectures in these frameworks. In our work we experience limitations in the ONNX file format and in the Unity Barracuda package, so we motivate further development of these. From C1 we uncover C5, stating that \textbf{we see a too marked separating line between the development-sided frameworks and the operation-sided ones}. C5 reveals a constraint we found in DL-based software defined as follows. \textbf{No matter what framework one wishes to use, it will always need to wrap at least two different technologies}: one for the data management and model training outside the device, and the other for the model operation inside the user’s device. This implies challenges regarding the mutual support between the two technologies. In the following paragraph, the faced challenges regarding C5 are listed.

First, \textbf{the pre-processing steps that are applied to both the initial dataset in the development phase and the ones in the operation phase (which are applied in real time to the images taken by the user) have to be the same despite being implemented in different frameworks}. DL-based software developers shall ensure that the inputs on both platforms are the same to the network, so it gives the same predictions for the same images.
Secondly, \textbf{the operation-sided framework must support model importing and must provide functional libraries that enable the appropriate usage of the model in its exported format}. In Unity Barracuda, a challenging limitation found is the inability to support a lot of the basic operations in the treatment of DL models (a reason for which we verify C1). For example, let's consider the \textit{view()} function from PyTorch. For the hand-made CNNs, this operation has been replaced by \textit{Flatten()} which does exactly the same, but for the pre-trained MBv2 this method is used and hence the architecture cannot be deployed directly to Unity. In conclusion, due to the limitations in the used operation-sided framework, a model architecture that is built for ensuring efficiency on mobile devices cannot be deployed to the mobile application.

Regarding C2, \textbf{we have experienced high volatility in the applications' effectiveness when testing them with real-world data outside of the training set}. We verify C2 because we experience that the DL-based mobile applications effectiveness strongly depends on the data that is fed to it. This fact makes sustainability (C4) \textbf{become a critical quality attribute for reducing the data dependency and ensuring high performance of the models when deployed}. In the practical study, we implement a solution for the challenge of ensuring sustainability in DL-based software that consists of real world context data augmentation. By doing this, we smooth the differences between the development and operation phases of the DL-based software lifecycle by adapting the models to the real world environment. Hence, we also smooth the dependency between the applications' effectiveness and the data that is fed to it, which makes the DL-based mobile applications more sustainable and makes them generalize better. See the performance of the two deployed models in the two iterations of the operation phase in Table \ref{tab:operation-performance} where it is shown how the data augmentation technique applied in Iteration 2 reduces the data-software dependency (C2) and hence increases the sustainability (C4) of the applications. 

\begin{table}[ht]
\centering
\caption{Performance of the deployed models when receiving operation-sided data}
\label{tab:operation-performance}
\scalebox{1}{%
    \begin{tabular}{|c|c|r|r|}
        \hline
         \textbf{Model} & \textbf{Iteration} & 
         \textbf{Accuracy} & \textbf{Storage Weight} \\
         \hline
         Small CNN & 1 & 45.07\% & 22.11MB \\
         RN34 & 1 & 64.78\% & 83.24MB \\
         Small CNN & 2 & \textbf{92.95\%} & 22.11MB \\
         RN34 & 2 & \textbf{98.59\%} & 83.24MB \\
         \hline
    \end{tabular}
}
\end{table}

Regarding C3, our approach for providing explainability in the built applications is based on providing standard confidence indicators of the models' predictions. These indicators consist of the output probabilities of class membership which have been scaled to become percentages of confidence for a given input. \textbf{These simple indicators provide a level of explainability of the models} that is acceptable for the applications in the practical study, meaning that \textbf{they allow some level of interpretation to the user}. We do not consider ensuring this quality attribute a critical challenge because it might not generalize to lots of applications that integrate a DL-based component. However, \textbf{we recognize the importance of the presence of any measure that provides the user with information about why the model has given a result}.

Our practical study uncovered C6. Compared to the usage of a server, the model exporting operation and its integration into the mobile application \textbf{has the drawback of the inclusion of the model's storage weight in the application, hence, in the mobile storage}. For this reason the \textbf{optimization of the performance trade-off is a must when deploying DL-based software to mobile devices}. We have shown in the practical study that not limiting the complexity of the model's architecture, as in the case of the RN34, yields a cost in performance in the mobile application in terms of real time response and storage weight. For the deployment of DL-based software to mobile devices, ensuring efficiency is not an option, it is a must, and facing it becomes a critical challenge.

\subsection{RQ2: Criteria needed to reason about the trade-off between accuracy and complexity of DL models in mobile applications}
Most of the criteria applied to optimize the performance trade-off are approaches based on the reduction of the models' complexity. In an environment where the developer has few alternatives in the choice of both development-sided and operation-sided frameworks, the deployment of these systems becomes a bottleneck of the DL pipeline. This way, \textbf{the evolution of the performance of DL-based software is strictly linked to the evolution of the current model conversion and exporting technologies}, a field that is in its very early stages, where few file formats can be used and with several limitations. For this reason, the deployment of DL-based software to the server cloud and the remote usage through internet connections is still a more flexible choice in terms of DL-based services that can be provided to a mobile user. However, this is not fault-tolerant, due to the dependency on hosting and internet connections. 

\textbf{A fact that makes the deployment of DL-based software to mobile devices not always suitable is that it collides with the benefits of transfer learning}. The major capabilities of transfer learning include great results with little task-specific data, the ease in fine-tuning pre-built architectures and omitting the model design stage. The common available models for transferring learning are of massive weight, and the revolution of transfer learning has created a motivation to use these largest-scaled models for solving any task before designing task-specific architectures \cite{transfer-learn}. This has never been a problem until now, when one wants the massive model to be deployed in an environment with limited computational power.

We show in Table \ref{tab:operation-performance} that the most famous architecture amongst all the experimented ones, which is the RN34 \cite{resnet}, is the one carrying the biggest drawbacks in terms of storage weight. In this way, the simple CNN architecture with the appropriate configuration outperforms the massive model in terms of performance because of the ability to provide almost the same accuracy when solving the traffic sign recognition task and because of its reduction in the complexity of the architecture.

Finally, the MBv2 looks very promising in this field for its low complexity and high accuracy. Although it has not been deployed for the proposed practical study due to framework incompatibilities, it is concluded that the modifications in its architecture provide the best capabilities for the purpose of deploying DL-based software into environments with limited computational power, in this case the mobile devices. 

This way, in order to lean towards a specific architecture for DL models that are going to be deployed in mobile applications we look for the ones that keep low complexity (e.g. fast real-time response, limited number of parameters, reasonable storage weight) and accomplish high accuracy and generalization. 

\section{Discussion}
In the following we review our findings and discuss their implications. First, we have verified that the identified challenges that regard the software-data dependency and sustainability of the models deployed in the software are of major concern. To solve them, we have proposed the implementation of a data augmentation technique that adapts the models to the real world environment, hence providing sustainability. We have shown significant increases in the accuracy when sustainability is ensured in the two systems by means of the data augmentation technique (see Table \ref{tab:operation-performance}). This step is represented by the "Send Data" operation in Figure \ref{global_overview}. We have also verified the need of the evolution of frameworks that support the development of DL-based software, which is another challenge that has been identified in related work. Additionally, we have uncovered a new challenge related to the mutual dependencies of these frameworks, which enriches the identification of the framework capabilities challenge. Furthermore, we have provided identification and description of the criteria that can be applied to approach efficiency in the deployment of DL models in mobile applications. Also, we have related the reasoning of the accomplishment of efficiency to a newly identified challenge.

Taking into account the differences between the performance of the two DL-based mobile applications according to our criteria, we now discuss a possible functional and efficient implementation. This implementation could consist of the integration of the two deployed models in the same DL-based mobile application. The default model could be set to the most efficient one, and in case the results given by this were not satisfactory the user would have the option to switch to the most robust and least efficient model.

\section{Conclusions}
In this work we have studied the DL-based software lifecycle in a practical manner. We have provided an analysis of the challenges found when developing this type of software. Specifically, we have put the focus on the deployment of DL-based software in mobile applications. Concretely, we have tested the performance of different CNN architectures under the PyTorch and Android frameworks. With all this, we have experienced, solved and documented many of the previously identified challenges in the related work and also have highlighted new ones. Furthermore, we have analysed the roles of the accuracy and the complexity in the DL-based software performance, relating this quality attribute to a newly identified challenge of the deployment of DL-based software.

This study motivates several key points of future work. First, the search for optimized model architectures that allow more efficient solutions to the deployment of DL-based components in mobile applications. These can both implement more efficient operations in its architecture or have limitations in its complexity when designed. The study also encourages the community to develop alternative solutions to the conversion and export of models together with software technologies that support mobile application development with DL-based components. The latter would reduce the severity of the challenges that are met while implementing this type of software. 
Moreover, the study motivates the design of more sophisticated confidence indicators that describe in a more precise way the uncertainty in the models predictions. The uncertainty of a model prediction given the input conditions is a measure that is hard to determine but useful to influence the decisions of autonomous DL-based systems. These can be key in many applications related to safety-critical environments. Also, sophisticated confidence indicators increase the quality of the systems since these enhance the explainability to the user. Last, we expect long evolution in the design of empowered GPUs for mobile devices which can relax the limitations of complexity in the architectures for achieving efficiency within the mobile device settings.  

\section{Data Availability}
In this section we provide a demo video of the developed and evolved DL-based software, available at \url{https://www.youtube.com/watch?v=yFsp6kxO5kI}. Additionally, we provide an open source code repository of the whole project in \url{https://github.com/yuyecreus/CNN-in-mobile-device}.

\section*{Acknowledgment}
We thank Lisa J{\"o}ckel for having reviewed our work and for being supportive and useful to enhance it.
The research presented in this paper has been developed in the context of the CBI course at the GCED@FIB.

{\small
\bibliographystyle{IEEE/IEEEtran}
\bibliography{references}
}

\end{document}